\def\year{2020}\relax
\documentclass[letterpaper]{article} 
\usepackage{aaai20}  
\usepackage{times}  
\usepackage{helvet} 
\usepackage{courier}  
\usepackage[hyphens]{url}  
\usepackage{graphicx} 
\usepackage{graphicx}  
\newcommand{\citet}[1]{\citeauthor{#1} \shortcite{#1}}
\newcommand{\citep}{\cite}

\title{BERT-based Acronym Disambiguation with Multiple Training Strategies}
\author{ \Large \textbf{Chunguang Pan \textsuperscript{\rm 1},Bingyan Song \textsuperscript{\rm 1}, Shengguang Wang \textsuperscript{\rm 1}, Zhipeng Luo \textsuperscript{\rm 1}}\\ 

\textsuperscript{\rm 1} DeepBlue Technology (Shanghai) Co., Ltd \\
\{panchg, songby, wangshg, luozp\} @deepblueai.com
}

\begin{document}

\maketitle

\begin{abstract}
Acronym disambiguation (AD) task aims to find the correct expansions of an ambiguous ancronym in a given sentence. Although it is convenient to use acronyms, sometimes they could be difficult to understand. Identifying the appropriate expansions of an acronym is a practical task in natural language processing. Since few works have been done for AD in scientific field, we propose a binary classification model incorporating BERT and several training strategies including dynamic negative sample selection, task adaptive pretraining, adversarial training and pseudo labeling in this paper. Experiments on SciAD show the effectiveness of our proposed model and our score ranks 1st in SDU@AAAI-21 shared task 2: Acronym Disambiguation. 
\end{abstract}

\section{Introduction}
An \textbf{acronym} is a word created from the initial components of a phrase or name, called the \textbf{expansion} \cite{jacobs2020acronyms}. In many literature and documents, especially in scientific and medical fields, the amount of acrnomys is increasing at an incredible rate. By using acronyms, people can avoid repeating frequently used long phrases.
For example, CNN is an acronym with the expansion \underline{C}onvolutional \underline{N}eural \underline{N}etwork, though it has additional expansion possibilities depending on context, such as \underline{C}ondensed \underline{N}earest \underline{N}eighbor. 

Understanding the correlation between acronyms and their expansions is critical for several applications in natural language processing, including text classification, question answering and so on.

Despite the convenience of using acronyms, sometimes they could be difficult to understand, especially for people who are not familiar with the specific area, such as in scientific or medical field. Therefore, it is necessary to develop a system that can automatically resovle the appropriate meaning of acronyms in different contextual information.

\begin{figure}[!t]
\centering
\includegraphics[width=3.2in]{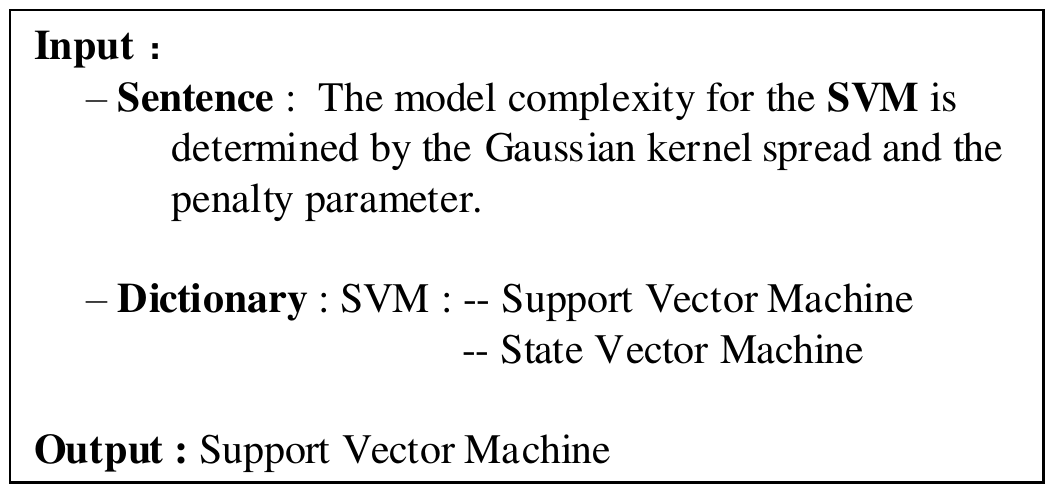}
\caption{An example of acronym disambiguation}
\label{fig_eg1}
\end{figure}

Given an acronym and several possible expansions, \textbf{acronym disambiguation(AD)} task is to determine which expansion is correct for a particular context. The scientific acronym disambiguation task is challenging due to the high ambiguity of acronyms. For example, as shown in Figure \ref{fig_eg1}, SVM has two expansions in the dictionary. According to the contextual information from the input sentence, the SVM here represents for the Support Vetor Machine which is quite smilar to State Vector Machine.

Consequently, AD is formulated as a classification problem, where given a sentence and an acronym, the goal is to predict the expansion of the acronym in a given candidate set. Over the past two decades, several kinds of approaches have been proposed. At the begining, pattern-matching techniques were popular. They \cite{taghva1999recognizing} designed rules and patterns to find the corresponding expansions of each acronym. However, as the pattern-matching methods require more human efforts on designing and tuning the rules and patterns, machine learning based methods (i.e. CRF and SVM) \cite{liu2017multi}  have been preferred. More recently, deep learning methods \cite{charbonnier2018using,jin2019deep} are adopted to solve this task. 

Recently, pre-trained language models such as ELMo \cite{peters2018deep} and BERT \cite{devlin2018bert}, have shown their effectiveness in contextual representation. Inspired by the pre-trained model, we propose a binary classification model that is capable of handling acronym disambiguation. We evaluate and verify the proposed method on the dataset released
by SDU@AAAI 2021 Shared Task: Acronym Disambiguation \cite{veyseh2020acronym}. Experimental results show that our model can effectively deal with the task and we win the first place of the competition.

\section{Related Work}
\subsection{Acronym Disambiguation}
Acronym diambiguation has received a lot of attentions in vertical domains especially in biomedical fields. Most of the proposed methods \cite{schwartz2002simple} utilize generic rules or text patterns to discover acronym expansions. These methods are usually under circumstances where acronyms are co-mentioned with the corresponding expansions in the same document. However, in scientific papers, this rarely happens. It is very common for people to define the acronyms somewhere and use them elsewhere. Thus, such methods cannot be used for acronym disambiguation in scientific field.

There have been a few works \cite{nadeau2005supervised} on automatically mining acronym expansions by leveraging Web data (e.g. click logs, query sessions). However, we cannot apply them directly to scientific data, since most data in scientific are raw text and therefore logs of the query sessions/clicks are rarely available.

\subsection{Pre-trained Models}
Substantial work has shown that pre-trained models (PTMs), on the large unlabeled corpus can learn universal language representations, which are beneficial for downstream NLP tasks and can avoid training a new model from scratch. 

The first-generation PTMs aim to learn good word embeddings. These models are usually very shallow for computational efficiencies, such as Skip-Gram \cite{mikolov2013distributed} and GloVe \cite{pennington2014glove}, because they themselves are no longer needed by downstream tasks. Although these pre-trained embeddings can capture semantic meanings of words, they fail to caputre higher-level concepts in context, such as polysemous disambiguation and semantic roles. The second-generation PTMs focus on learning contextual word embeddings, such as ELMo \cite{peters2018deep}, OpenAI GPT \cite{radford2018improving} and BERT \cite{devlin2018bert}. These learned encoders are still needed to generate word embeddings in context when being used in downstream tasks.

\subsection{Adversarial Training}
Adversarial training (AT) \cite{goodfellow2014explaining} is a mean of regularizing classification algorithms by generating adversarial noise to the training data. It was first introduced in image classification tasks where the input data is continuous. 

\citet{miyato2016adversarial} extend adversarial and virtual adversarial training to the text classification by applying perturbation to the word embeddings and propose an end-to-end way of data perturbation by utilizing the gradient information. \citet{zhu2019adversarial} propose an adversarial attention network for the task of multi-dimensional emotion regression, which automatically rates multiple emotion dimension scores for an input text.  

There are also other works for regularizing classifiers by adding random noise to the data, such as dropout \cite{srivastava2014dropout} and its variant for NLP tasks, word dropout \cite{iyyer2015deep}. \citet{xie2019data} discusses various data noising techniques for language models and provides empirical analysis validating the relationship between nosing and smoothing. \citet{sogaard2013part} and \citet{li2017robust} focus on linguistic adversaries.

Combining multiple advantages in above works, we propose a binary classification model utilizing BERT and several training strategies such as adversarial training and so on.

\section{Data}
In this paper, we use the AD dataset called \textbf{SciAD} released by \citet{veyseh2020does}. They collect a corpus of 6,786 English papers from arXiv and these papers consist of 2,031,592 sentences that could be used for data annotation. 

\begin{figure}[!t]
\centering
\includegraphics[width=3.3in]{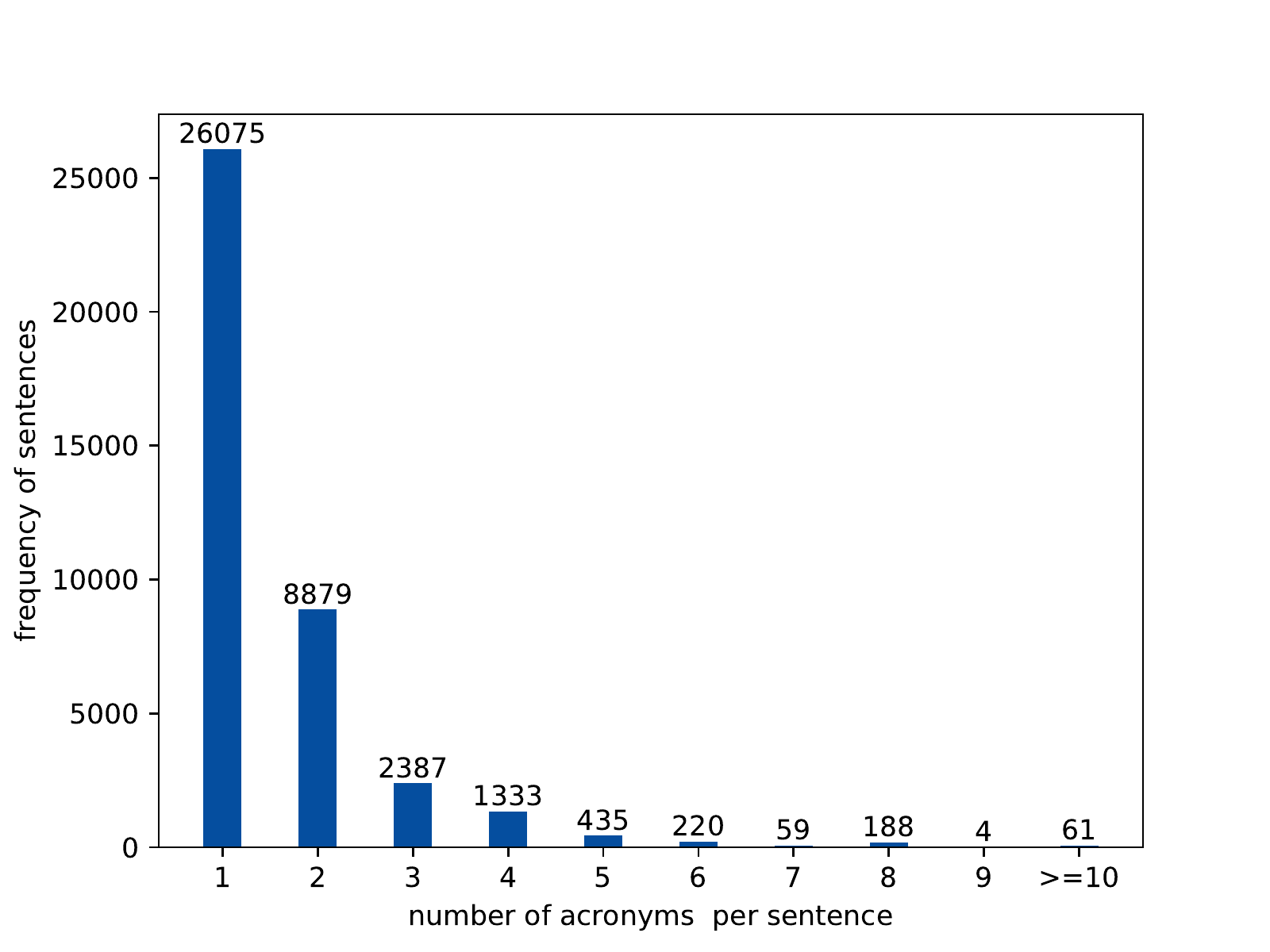}
\caption{Number of acronyms per sentence} 
\label{fig_data1}
\end{figure}

\begin{figure}[!t]
\centering
\includegraphics[width=3.3in]{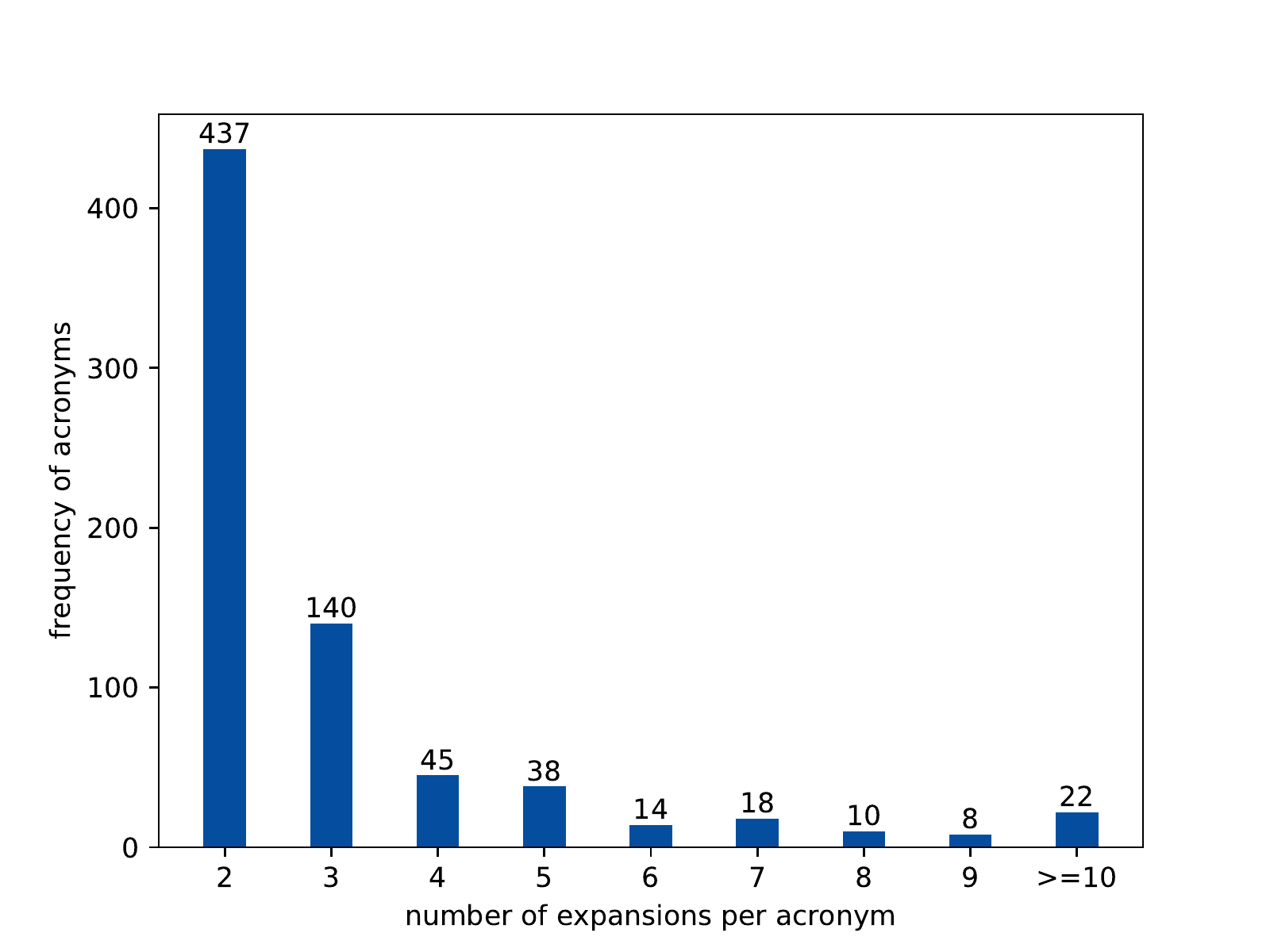}
\caption{Number of expansions per acronym} 
\label{fig_data2}
\end{figure}

\begin{figure*}[hpt]
    \centering
    \includegraphics[width=2.1\columnwidth]{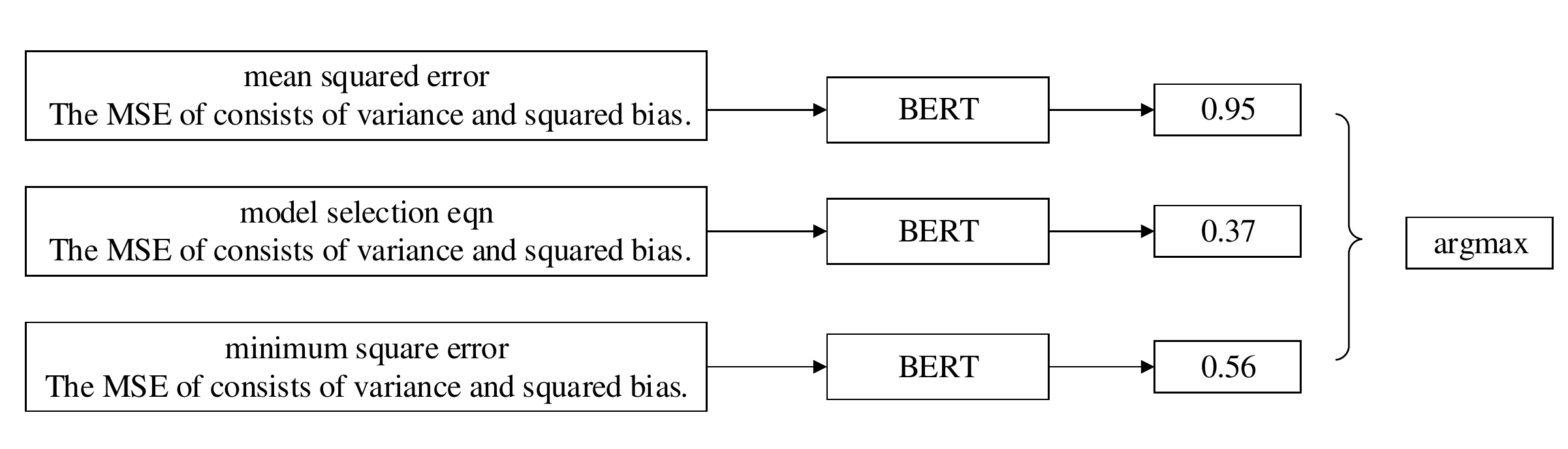} 
    \caption{Acronym disambiguation based on binary classification model. For each sample, the model needs to predict whether the given expansions matches the acronym or not, and find the expansion with the highest score as the correct one. }
    \label{binary_model0}
\end{figure*}

The dataset contains 62,441 samples where each sample involves a sentence, an ambiguous acronym, and its correct meaning (one of the meanings of the acronym recorded by the dictionary , as shown in \ref{fig_eg1}).

Figure \ref{fig_data1} and Figure \ref{fig_data2} demonstrate statistics of SciAD dataset. More specifically, Figure \ref{fig_data1} reveals the distribution of number of acronyms per sentence. Each sentence could have more than one acronym and most sentences have 1 or 2 acronyms. Figure \ref{fig_data2} shows the distribution of number of expansions per acronym. The distribution shown in this figure is consistent with the same distribution presented in the prior work (Charbonnier and Wartena, 2018) in which in both distributions, acronyms with 2 or 3 meanings have the highest number of samples in
the dataset \cite{veyseh2020does}.

\section{Binary Classification Model}
The input of the binary classification model is a sentence with an ambiguous acronym and a possible expansion. The model needs to predict whether the expansion is the corresponding expansion of the given acronym. Given an input sentence, the model will assign a predicted score to each candidate expansion. The candidate expansion with the highest score will be the model output. Figure \ref{binary_model0} shows an example of the procedure.

\subsection{Input Format}
Since BERT can process multiple input sentences with segment embeddings, we use the candidate expansion as the first input segment, and the given text as the second input segment. We separat these two input segments with the special token \begin{tt}[CLS]\end{tt}. Furthermore, we add two special tokens \begin{tt}<start>\end{tt} and \begin{tt}<end>\end{tt} to wrap the acronym in the text, which enables that the acronym can get enough attention from the model. 

\subsection{Binary Model Architecture}
\begin{figure}[htp]
    \centering
    \includegraphics[width=0.99\columnwidth]{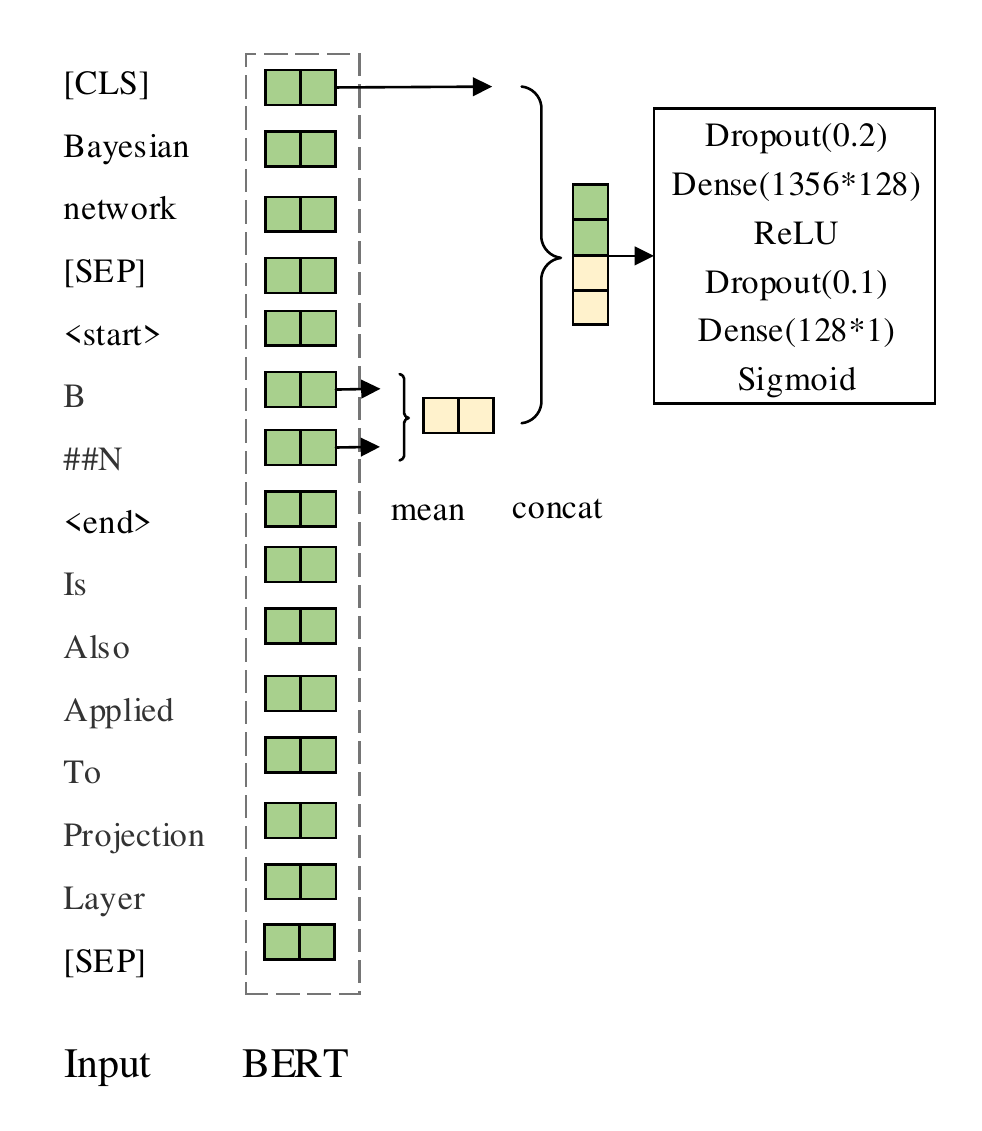} 
    \caption{The binary classification model. }
    \label{binary_model1}
\end{figure}

The model architecture is described in Figure \ref{binary_model1} in detail. First, we use a BERT encoder to get the representation of input segments. Next, we calculate the mean of the start and end positions of the acronym, and concatenate the representation with the \begin{tt}[CLS]\end{tt} position vector. Then, we sent this concatenated vector into a binary classifier for prediction. The represenation first pass through a dropout layer \citep{srivastava2014dropout} and a feedforward layer. The output of these layers is then feed into a ReLU \citep{glorot2011deep} activation. After this, the calculated vector pass through a dropout layer and a feedforward layer again. The final prediction can be obtained through a sigmoid activation.

\subsection{Training Strategies}
\subsubsection{Pretrained Models}
Experiments from previous work have shown the effectiveness of pretrained models. Starting from BERT model, there are many improved pretrained models. Roberta uses dynamic masks and removes next sentence prediction task. In our experiments, we compare BERT and Roberta models trained on corpus from different fields. 

\subsubsection{Dynamic Negative Sample Selection 
\label{negative:sample}}  During training, we dynamicly select a fixed number of negative samples for each batch, which ensures that the model is trained on more balanced positive and negative data, and all negative samples are used in training at the same time.

\subsubsection{Task Adaptive Pretraining} 
\citet{gururangan-etal-2020-dont} shows that task-adaptive pretraining (TAPT) can effectively improve model performance. The task-specific dataset usually covers only a subset of data used for general pretraining, thus we can achieve significant improvement by pretraining the masked language model task on the given dataset.

\subsubsection{Adversarial Training} 
Adversarial training is a popular approach to increasing robustness of neural networks. As shown in \citet{miyato2016adversarial}, adversarial training has good regularization performance. By adding perturbations to the embedding layer, we can get more stable word representations and a more generalized model, which significantly improves model performance on unseen data. 

\subsubsection{Pseudo-Labeling}
Pseudo labeling \citep{iscen2019label,oliver2018realistic,shi2018transductive} uses network predictions with high confidence as labels. We mix these pseudo labels and the training set together to generate a new dataset. We than use this new dataset to train a new binary classification model. Pseudo-labeling has been proved an effective approach to utilize unlabeled data for a better performance.

\section{Experiments}
\subsection{Hyper parameters} 
The batch size used in our experiments is 32. We train each model for 15 epochs. The initial learning rate for the text encoder is $1.0 \times 10^{-5}$, and for other parameters, the initial learning rate is set to $5.0 \times 10^{-4}$. We evaluate our model on the validation set at each epoch. If the macro F1 score doesn't increase, we then decay the learning rate by a factor of 0.1. The minimum learning rate is $5.0 \times 10^{-7}$. We use Adam optimizer \cite{kingma2017adam} in all our experiments. 

\subsection{Pretrained Models} 
Since different pretrained models are trained using different data, we do experiments on several pretrained models. Table \ref{pretrained_table} shows our experimental results on different pretrained models in validation set. The bert-base model gets the highest score in commonly used pretrained models (the top 3 lines in Table \ref{pretrained_table}). Since a large ratio of texts in the given dataset come from computer science field, the cs-roberta model outperforms the bert-base model by 1.6 percents. The best model in our experiments is the scibert model, which achieves the F1 score of $89\%$. 
\begin{table}[htp]
    \centering
\begin{tabular}{l|c c c c}
    Model  & Precision &  Recall & F1 \\
    \hline
    bert-base-uncased  & $0.9176$ & $0.8160$ & $0.8638$ \\
    bert-large-uncased  &$0.9034$ & $0.7693$ & $0.8311$ \\
    roberta-base & $0.9008$ & $0.7687$ & $0.8295$ \\
    cs-roberta-base  & $0.9216$ & $0.8415$ & $0.8797$ \\
    scibert-scivocab-uncased  & $\textbf{0.9263}$ & $\textbf{0.8569}$ & $\textbf{0.8902}$ \\
\end{tabular}
\caption{Results on validation set using different pretrained models. }
\label{pretrained_table}
\end{table}

\subsection{Training Procedure}
We incorporate all the training strategies introduced above to improve the performance of our proposed binary classification model. According to the experiment result in Table \ref{pretrained_table}, we choose scibert as the fundamental pretrained model and use the TAPT technique to train a new pretrained model. Then we add the dynamic negative sample selection and adversarial training strategies to train the binary classfication model. After this, we utilize the pseudo-labeling technique and obtain the final binary classification model.

\subsection{Further Experiments}
\subsubsection{Combining training strategies} 
We do some futher experiments on validation set to verify the effectiveness of each strategy mentioned above. The results are shown in Table \ref{training_table}.  As shown in the table, F1 score increases by 4 percents with dynamic sampling. TAPT and adversarial training further improve the performance on validation set by 0.47 percent. Finally, we use pseudo-labeling method. Samples from the test set with a score higher than 0.95 are selected and mixed with the training set. It still slightly improves the F1 score. 

\begin{table}[htp]
    \centering
\begin{tabular}{l|c c c c}
    Model  & Precision & Recall &  F1 \\
    \hline
    scibert-scivocab-uncased & $0.9263$ & $0.8569$ & $0.8902$ \\
    +dynamic sampling   & $0.9575$ & $0.9060$ & $0.9310$ \\
    +task adaptive pretraining  & $0.9610$ & $0.9055$ & $0.9324$ \\
    +adversarial training  & $\textbf{0.9651}$ & $0.9082$ & $0.9358$ \\
    +pseudo-labeling & $0.9629$ & $\textbf{0.9106}$ & $\textbf{0.9360}$ \\
\end{tabular}
\caption{Results on validation set using different training approaches. }
\label{training_table}
\end{table}

\subsubsection{Error Analysis}
We gather a sample of 100 development set examples that our model misclassified and look at these examples manually to do the error analysis. 

From these examples, we find that there are two main cases where the model gives the wrong prediction. The first one is that the candidate expansions are too similar, even have the same meanings in different forms. For example, in the sentence 'The SC is decreasing for increasing values of ...', the correct expansion for 'SC' is 'sum capacities' while our prediction is 'sum capacity' which has the same meaning with the correct one but in the singular form. 

The second one is that there is too little contextual information in the given sentence for prediction. For instance, the correct expansion for 'ML' in sentence 'ML models are usually much more complex, see Figure.' is 'model logic', the predict expansion is 'machine learning'. Even people can hardly tell which one is right only based on the given sentence. 

\subsubsection{Time complexity}
To analysis the time complexity of our proposed method, we show measurements of the actual running time observed in our experiments. The discussions are not that precise or exhaustive. However, we believe they are enough to offer readers rough estimations of the time complexity of our model.

We utilize TAPT strategy to further train the scibert model by using eight NVIDIA TITAN V (12GB). It takes three hours to train 100 epochs in total.

After getting the new pretrained model, we trained the binary classification model on two NVIDIA TITAN V. On average, each epoch of the training and inference time of adding adversarial training and pseudo-labeling are shown in Table \ref{Time complexity} respectively. It begins to converge after five epochs. It takes nearly the same time to do the inference while the training time is twice as long after adversarial training is added.
\begin{table}[htp]
    \centering
\begin{tabular}{l|c c c c}
    Model  & Train & Inference \\
    \hline
      & $1588s$ & $150.42s$\\
    +adversarial training  & $3021s$ & $149.64s$ \\
    +pseudo-labeling & $3328s$ & $149.36s$  \\
\end{tabular}
\caption{Time complexity }
\label{Time complexity}
\end{table}

\subsubsection{Comparison Results}
We compared our results with several other models. Precision, Recall and F1 of our proposed model are computed on testing data via the cross-validation method.
\begin{itemize} 
\item \textbf{MF \& ADE} Non-deep learning models that utilize rules or hand crafted features \cite{li2018guess}.
\item \textbf{NOA \& UAD} Language-model-based baselines that train the word embeddings using the training corpus \cite{charbonnier2018using,ciosici2019abbreviation}.
\item \textbf{BEM \& DECBAE} Models employ deep architectures (e.g., LSTM) \cite{jin2019deep,blevins2020moving}.
\item \textbf{GAD} A deep learning model utilizes the syntactical structure of the sentence \cite{veyseh2020does}.
\end{itemize}

\begin{table}[htp]
    \centering
\begin{tabular}{l|c c c c}
    Model & Precision & Recall & F1 \\
    \hline
    MF & $0.8903$ & $0.4220$ & $0.5726$ \\
    ADE & $0.8674$ & $0.4325$ & $0.5772$  \\
    NOA & $0.7814$ & $0.3506$ & $0.4840$ \\
    UAD & $0.8901$ & $0.7008$ & $0.7837$  \\
    BEM & $0.8675$ & $0.3594$ & $0.5082$ \\ 
    DECBAE & $0.8867$ & $0.7432$ & $0.8086$\\ 
    GAD & $0.8927$ & $0.7666$ & $0.8190$  \\ 
    \textbf{Ours} & $\textbf{0.9695}$ & $\textbf{0.9132}$  & $\textbf{0.9405}$  \\
    Human Performance & $0.9782$ & $0.9445$ & $0.9610$ \\ 
\end{tabular}
\caption{Results of different models on testing dataset}
\label{test_table}
\end{table}

As shown in Table \ref{test_table}, rules/features fail to caputre all patterns of expressing the meanings of the acronym, resulting in poorer recall on expansions compared to acronyms. In contrast, the deep learning model has comparable recall on expansions and acronyms, showing the importance of pre-trained word embeddings and deep architectures for AD. However, they all fall far behind human level performance. Among all the models, our proposed model achieves the best results on the SciAD and is very close to the human performance which shows the capability of the strategies we introduced above.

\subsubsection{SDU@AAAI 2021 Shared Task: Acronym Disambiguation}
The competition results are shown in Table \ref{ranking}. We show scores of the top 5 ranked models as well as the baseline model. The baseline model is released by the provider of the SciAD dataset \cite{veyseh2020does}. Our model performs best among all the ranking list and outperforms the second place by $0.32\%$. In addition, our model outperforms the baseline model by $12.15\%$ which is a great improvement.
\begin{table}[htp]
    \centering
\begin{tabular}{l|c c c c}
    Model &  Precision &Recall &  F1 \\
    \hline
    \textbf{Rank1} & $\textbf{0.9695}$ & $\textbf{0.9132}$  & $\textbf{0.9405}$  \\
    Rank2 & $0.9694$ & $0.9073$ & $0.9373$  \\
    Rank3 & $0.9652$ & $0.9009$ & $0.9319$ \\
    Rank4 & $0.9595$ & $0.8959$ & $0.9266$  \\
    Rank5 & $0.9548$ & $0.8907$ & $0.9216$ \\ 
    Baseline & $0.8927$ & $0.7666$ & $0.8190$  \\ 
\end{tabular}
\caption{Leaderboard}
\label{ranking}
\end{table}

\section{Conclusion}
In this paper, we introduce a binary classification model for acronym disambiguation. We utilize the BERT encoder to get the input representations and adopt several strategies including dynamic negative sample selection, task adaptive pretraining, adversarial training and pseudo-labeling. Experiments on SciAD show the validity of our proposed model and we win the first place of the SDU@AAAI-2021 Shared task 2. 

\bibliographystyle{aaai}
\bibliography{citations}

\end{document}